\documentclass{article}

\PassOptionsToPackage{numbers, compress}{natbib}
\usepackage[final]{tackling_climate_workshop_style}

\usepackage[utf8]{inputenc} 
\usepackage[T1]{fontenc}    
\usepackage{hyperref}       
\usepackage{url}            
\usepackage{booktabs}       
\usepackage{amsfonts}       
\usepackage{nicefrac}       
\usepackage{microtype}      
\usepackage{tabularx}
\usepackage{multirow}
\usepackage{mathtools}
\usepackage{tabularx}
\usepackage{xcolor,colortbl}
\usepackage{nicematrix}
\usepackage[export]{adjustbox}
\usepackage{amsmath}
\usepackage[ruled,linesnumbered]{algorithm2e}
\usepackage{wrapfig}
\usepackage{tikz}

\title{Self-Supervised Pre-Training for Precipitation Post-Processor}

\author{%
  Sojung An\thanks{These authors contributed equally to this work.} \\
  KIAPS\\
  Seoul, Republic of Korea \\
  \texttt{sojungan@kiaps.org} \\
  \And
  Junha Lee\footnotemark[1] \\
  KITECH\\
  Ansan, Republic of Korea \\
  \texttt{junha@kitech.re.kr} \\
  \And
  Jiyeon Jang\\
  KIAPS\\
  Seoul, Republic of Korea \\
  \texttt{jyjang@kiaps.org} \\
  \AND
  Inchae Na\\
  KIAPS\\
  Seoul, Republic of Korea \\
  \texttt{icna@kiaps.org} \\
  \And
  Wooyeon Park \\
  KIAPS \\
  Seoul, Republic of Korea \\
  \texttt{wooyeon@kiaps.org} \\
  \And
  Sujeong You \\
  KITECH\\
  Ansan, Republic of Korea \\
  \texttt{sjyou21@kitech.re.kr}
}

\begin{document}

\maketitle

\begin{abstract}
Obtaining a sufficient forecast lead time for local precipitation is essential in preventing hazardous weather events.
Global warming-induced climate change increases the challenge of accurately predicting severe precipitation events, such as heavy rainfall.
In this paper, we propose a deep learning-based precipitation post-processor for numerical weather prediction (NWP) models.
The precipitation post-processor consists of (i) employing self-supervised pre-training, where the parameters of the encoder are pre-trained on the reconstruction of the masked variables of the atmospheric physics domain; and (ii) conducting transfer learning on precipitation segmentation tasks (the target domain) from the pre-trained encoder.
In addition, we introduced a heuristic labeling approach to effectively train class-imbalanced datasets.
Our experiments on precipitation correction for regional NWP show that the proposed method outperforms other approaches.
\end{abstract}

\section{Introduction}

In modern society, precipitation forecasting plays a vital role in the response to and prevention of social and economic damage.
Deep learning has been rapidly utilized in precipitation forecasting, such as for simulating echo movements and predicting typhoon trajectories based on observational data \cite{shi:2015, shi:2017, ravuri:2019, an:2023}.
However, exclusively relying on observational data fails to capture the fundamental physical and dynamic mechanisms in the real world. This results in an exponential increase in error as the forecast lead time increases \cite{espeholt:2022}.
In addition, climate change increases the uncertainty of predicting extreme events such as torrential rains \cite{nissen:2017}.
A limited forecast lead time makes it difficult to prepare  for extreme weather events in advance.

Recent research has been actively conducted to enhance forecast lead times via the post-processing of numerical weather prediction (NWP) model  data \cite{kim:2022, ghazvinian:2021, rojas:2023, zhang:2021}.
Espeholt et al.~\cite{espeholt:2022} proposed Metnet2, a 12-h probabilistic forecasting model. They designed a hybrid model consisting of a forecasting model based on observational data and post-processing NWP model data.
While Metnet2 displayed promising results, simulating heavy rainfall remains a challenging task.

Therefore, we designed a self-supervised pre-training process that considers the physical and dynamic processes among atmospheric variables to improve the relative bias and reliability of predicting heavy rain.
To this end, we randomly masked three-dimensional (3D) images \cite{feichtenhofer:2022} and trained an encoder-decoder to reconstruct variables based on dynamic sparse kernels (DSKs) \cite{wang:2023}. 
Subsequently, we trained a decoder for probability-based precipitation correction based on a pre-trained encoder. 
The pre-training process helps in conducting a correlation analysis based on the NWP model variables and provides additional data-augmentation effects.
Finally, we propose  a continuous labeling method for learning class-imbalanced datasets. This method facilitates a continuous probability distribution based on the precipitation density.

\begin{figure}
\centering
\includegraphics[width=.95\linewidth, height=4.5cm]{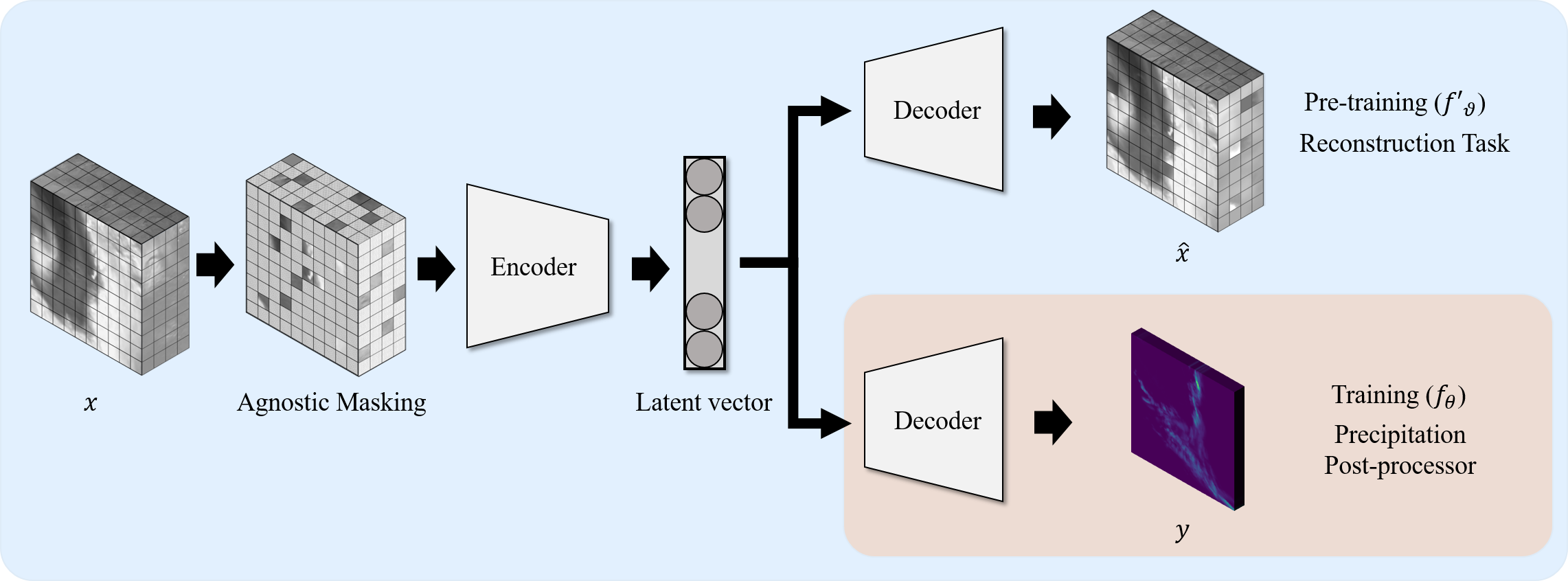}
\caption{Process of learning the precipitation post-processor. 
The model consists of two main phases: 1) pre-training the encoder and decoder using a reconstruction task after masking the inputs and 2) training the decoder for precipitation prediction using the trained encoder.
During the main training, the latent vector learned during pre-training is used as an encoder with fixed weights.}
\label{fig:en}
\end{figure}

\section{Method}
This section describes the proposed approach.
We first present a self-supervised pre-training procedure that aims to define the reconstruction task.
Subsequently, we describe the target process for the precipitation post-processing.

\subsection{Problem formulation}
Given a set of ground-truth pairs \textbf{G} = ($\left\{x_i\right\}_{i=1}^{n}, \left\{y_j\right\}_{j=1}^{m})$ with $x \in \mathbb{R}^{n \times h \times w}$, $y \in \mathbb{R}^{m \times h \times w}$, and $z \in \mathbb{R}^{d}$, let $x =\left\{x_1, \cdots, x_n\right\}^{\mathrm{T}}$ represent the input, and $y = \left\{y_1, \cdots y_m \right\}^{\mathrm{T}}$ represent the output. 
Our objective is to derive a precipitation segmentation function $f: \mathbb{R}^{d} \rightarrow \mathbb{R}^{c}$, assuming $z \in \mathbb{R}^{d}$ as a latent space with distribution $p(z)$ defined by the pre-training process.
The task involves reconstructing $f'_{\vartheta}$ from the NWP data to predict the target object function $f_{\theta}$.
The probabilistic encoder-decoder of the proposed method is defined as follows:
\begin{equation}
\mathcal{L}(\theta, \vartheta; x) = -\frac{1}{2}\mathbb{E}_{q_{\vartheta}(z|x)}[||x-\hat{x}||_{2}^2] + \mathrm{KL} (q_{\vartheta}(z|x)| p^{*}(y)),
\label{eq:object_function}
\end{equation}
where $\hat{x}$ denotes the reconstructed output from $x$ as generated by decoder, $q_{\vartheta}$ represents the pre-training process distribution, and $p^{*}$ is the class probability distribution, detailed in Section \ref{sec:cl}.
The loss function combines the mean squared error for minimizing pre-training loss and the Kullback-Leibler (KL) divergence term to quantify the difference between the learned latent variable distribution $q_{\vartheta}(z|x)$ and the class probability distribution $p^{*}(y)$.

\subsection{Continuous labeling}
\label{sec:cl}
In multi-class classification, instances $x$ are classified into one of the $y$ labels based on the ranges of precipitation intensity.
Let $p^{*}(y)=[\mathbb{P}(y|z)]$ be the class probability distribution and $\mathcal{L}$ be the cross-entropy loss. For the predictive function $f(x)$ and the label $y$, the cross-entropy loss $\mathcal{L}(f(x),y)$ is computed as follows:
\begin{align}
\begin{split}
   \mathcal{L}(f(x), y) = -\sum_{j=1}^{m} w_j \cdot p^{*}(y_{j})log(f(x)_{j}),
\end{split}
\end{align}
where $w_j$ denotes the weight set according to each class label $y_j$, and $p^{*}(\cdot)$ denotes the continuous labeling.
The minimization leads to the maximum likelihood estimate of the classifier parameters.
Minimization  yields a maximum likelihood estimate of the classifier parameters.
Algorithm \ref{alg1} presents the method for smoothing the probability values based on the density of the label range. This method differs from that of one-hot labeling, where the probability takes a value of one.
In the smoothing method, the sum of the probability values per class is fixed at one. 

\begin{algorithm}
\DontPrintSemicolon
\caption{Continuous labeling of the density of the rainfall range}\label{alg1}   
\KwIn{
  QPE $y = [0, 100)$; 
  rainfall threshold set $\gamma=\left\{r_0, \cdots, r_{m-2}\right\},\ (m \geq 2)$
 }
\KwOut {Probability label $\hat{Y}$}
  Initialize the number of rainfall thresholds $\gamma$ \\
  $p(y_j)= 
  \begin{cases}
  1,& \text{\small if $r_{j-1} \leq y < r_{j}$} \quad \tcc{\scriptsize Set the probability of label $k$ to 1.} \\
  0,& \text{\small otherwise}  \quad \tcc{\scriptsize Set the probability of label $i$ to 0 if outside the range.}\\
  \end{cases}
  $\;
\While{ $j$\ \ }{ 
  \Switch{$ p(y_j) $}{
    \textbf{if} $\{x \leq r_{j-1}\} \quad p^{*}(y_j) \rightarrow 0$ \\
    \textbf{else if} $\{r_{j-1} < y_j \leq r_{j}\} \quad p^{*}(y_j) \rightarrow \frac{r_{j+1}-y_j}{r_{j+1} - r_{j}}$\\
    \textbf{else if} $\{r_{j} < y_j \leq r_{j+1}\} \quad p^{*}(y_j) \rightarrow 1-\frac{y_j-r_{j}}{r_{j} - r_{j-1}}$\\
    \textbf{else if} $\{j=m-2\} \quad p^{*}(y) \rightarrow 1$\\
    \textbf{else} $\Longrightarrow  \quad p^{*}(y_j) \rightarrow 0$
    }
}
    \Return{$p^{*}(y)$}
\end{algorithm}
Given a rainfall $y_c$ that lies between two thresholds $r_{j}$ and $r_{j+1}$, assume that $y_c$ is closer to $r_{j+1}$.
The probability values for each threshold are sets $p(y_{c=j})=1$ and $p(y_{c\neq j})=0$. 
In this context, $p(\cdot)$ denotes the probability value based on the original labeling. 
According to Algorithm \ref{alg1}, both $p^{*}(y_{j})$ and $p^{*}(y_{j+1})$ are non-zero, $p^{*}(y_c)$ has a probability value of $p^{*}(y_j)<p^{*}(y_{j+1})$, and $p^{*}(y_j)+p^{*}(y_{j+1})=1$.
Setting the probability value using one-hot labeling can be a limitation in learning the uncertainty of the NWP. 
By continuously smoothing the probability value of a label, the method can help reduce uncertainty about the likelihood of being a different label.

\subsection{Training model}
This section presents our post-processing method that perturbs an entire patch for the reconstruction task. 
This method uses InternImage \cite{wang:2023} as an encoder and UPerNet \cite{xiao:2018} as a decoder.

\textbf{Patch embedding.}
We tokenize the input data into nonoverlapping spatial–temporal patches \cite{fan:2021}. 
Each 3D patch has dimensions $\mathcal{M} \in \mathbb{R}^{t \times p \times p}$, where t and p denote token size of time and height/width respectively.
This patching approach results in $x \in \mathbb{R}^{\frac{n}{t} \times \frac{h}{p} \times \frac{w}{p}}$ tokens. 
We then flatten the data and transform the tokens into $x \in \mathbb{R}^{\frac{nhw}{4p^{2}} \times \mathcal{C}}$ using a projection process. 
The data is added spatio-temporal positional encodings of the same size.

\textbf{Masking.}
Based on reference \cite{feichtenhofer:2022}, we use a structure-agnostic sample strategy to randomly mask the patches without using replacements from the set of embedded patches.
For a pre-training model in a reconstruction task, the optimal masking ratio is related to the amount of redundant information in the data \cite{devlin:2018, he:2022}.
Numerical forecasting models have a similar information redundancy, as each weather variable is at the same point in time. However, each variable has its own physical information.
Based on this, we use empirical results to determine the masking ratio.
The masking ratios of the pre-training and training are set to 90\% and 25\%, respectively.
The masked patches underwent layer normalization and are restored to their original input dimensions of $x' \in \mathbb{R}^{n \times h \times w}$.

\textbf{Encoder and decoder.}
We utilize DSK layers in the InternImage \cite{wang:2023} encoder for adaptive spatial aggregation.
InternImage is a backbone model that employs various techniques, in addition to attention, in the receptive field required for downstream tasks.
InternImage applies each encoder layer $e_i \subset e$ to a hierarchy of hidden states and the weights of each layer are stacked. 
We use the UperNet decoder to enable effective segmentation while preserving object boundaries and details from the training data.

\begin{table}[tb]
  \caption{Summary of results for our tasks. $\mathcal{I}$, $p^{*}$, and $\vartheta$ denote InternImage, label smoothing, and transfer learning, respectively. The tables compare the results of the models with thresholds of 0.1 and 10 mm.}
  \label{tab:result}
  \begin{tabular}{c|*{4}{c}|*{4}{c}}
    \hline
    \multirow{2}{*}{Method} & \multicolumn{4}{c|}{0.1 mm $\uparrow$} & \multicolumn{4}{c}{10 mm $\uparrow$} \\ 
    \cline{2-9}    
    & CSI & F1 & Precision & Recall & CSI  & F1 & Precision & Recall \\
    \hline
    RDAPS & 0.296 & 0.456 & 0.405 & 0.522 & 0.062 & 0.117 & 0.077 & \cellcolor{blue!15}\textbf{0.238}\\
    Metnet & 0.276 & 0.433 & 0.427 & 0.439 & 0.007 & 0.015 & 0.015 & 0.274\\
    Metnet+$p^{*}$ & 0.271 & 0.427 & 0.390 & 0.472 & 0.057 & 0.107 & 0.002 & 0.275\\
    $\mathcal{I}$+$p^{*}$ & 0.316 & 0.488 & 0.488 & 0.502 & 0.017 & 0.020 & 0.020 & 0.132\\
    $\mathcal{I}$+$\vartheta$ & 0.329 & 0.443 & 0.443 & \cellcolor{blue!15}\textbf{0.523} & 0.060 & 0.076 & 0.075 & 0.222 \\
    $\mathcal{I}$+$p^{*}$+$\vartheta$ (ours) & \cellcolor{blue!15}\textbf{0.347}& \cellcolor{blue!15}\textbf{0.515} & \cellcolor{blue!15}\textbf{0.620} & 0.481 & \cellcolor{blue!15}\textbf{0.093}  & \cellcolor{blue!15}\textbf{0.169} &  \cellcolor{blue!15}\textbf{0.130} & 0.227\\
    \hline
  \end{tabular}
\end{table}
\section{Experiments}

\begin{figure}[htb]
\centering
\begin{tabular}{c@{\hspace{-0.1pt}}p{1pt}c}
\textbf{\small (a) Original data \quad} & & 
\textbf{\small (b) Smoothed data \quad}\\
\includegraphics[width=0.44\textwidth]{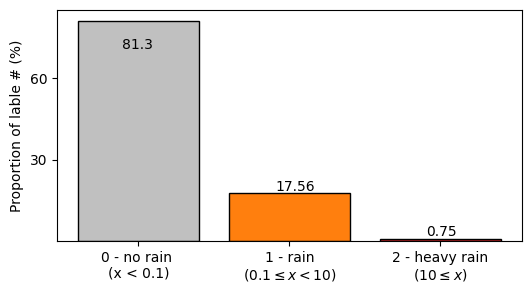} & \raisebox{15\height}{$\rightarrow$} &
\includegraphics[width=0.44\textwidth]{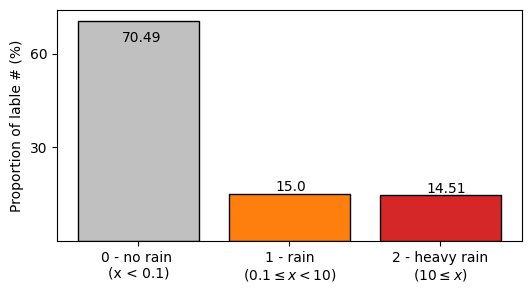}\\
\end{tabular}
\caption{Visualization of the proportion of labels in the training dataset: (a) is the original data and (b) shows the proportion for pixels with a non-zero probability smoothed using the method proposed in Section \ref{sec:cl}.}
\label{fig:proportion}
\end{figure}

We classified rainfall up to [0, 0.1) as no rain, [0.1, 10) as rain, and above 10 as heavy rain.
Figure \ref{fig:proportion} (a) shows the proportions of each label. 
The proportion of heavy rain within the training dataset was approximately 0.75\%. This indicates a significant data imbalance. The class imbalance issue frequently observed in precipitation measurement data is a common challenge in precipitation forecasting. Therefore, we investigated how the proposed method addresses class imbalance issues during the model training process.
We compared and evaluated our post-processor and heuristic label approaches with the state-of-the-art precipitation forecasting model called Metnet.

\begin{figure}[htb]
\centering
\includegraphics[width=0.98\textwidth]{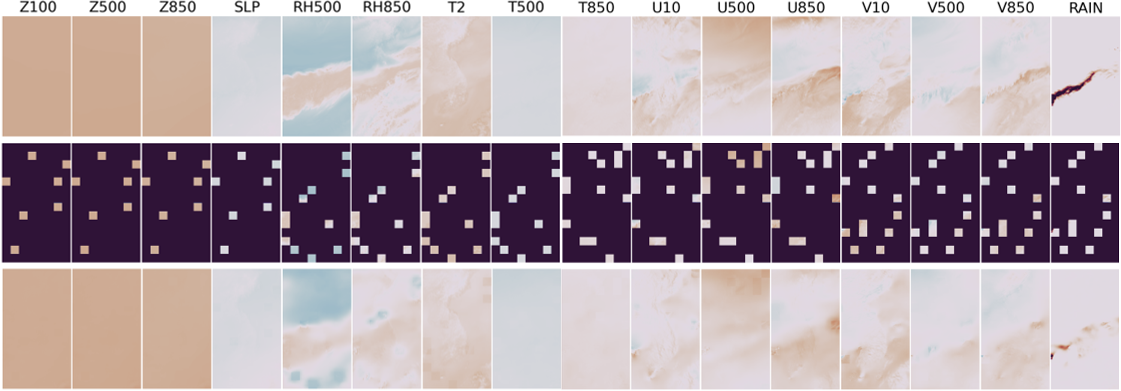}
\caption{Variable reconstruction results using the pre-trained model on data from August 10, 2022 at 00 UTC. The first row visualizes the normalized variables. The second row visualizes the variables with 90\% of the pixels masked. The third row shows the results of reconstructing the masked pixels. For the visualization, the masked values were set to -100, and a range of (-10, 10) was used. The number beside the variable indicates the vertical level.}
\label{fig:mask}
\end{figure}
Our main results are summarized in Table \ref{tab:result}. 
For the pre-trained model, we used a mask ratio of 90\%, a learning rate of 1.6e-3, and approximately 150 k iterations. For the training model, we used a mask ratio of 25\%, a learning rate of 1e-4, and approximately 35 k iterations.
Training the model using continuous labeling had a noticeable impact on the performance improvement.
We observed a significant enhancement in the accuracy of heavy rain when employing continuous labeling in Metnet.
We aimed to address (i) solving the imbalanced label problem and (ii) learning the weights between the variables in the self-supervised learning. 

Figure \ref{fig:mask} shows the reconstruction of 16 masked variables. 
Using the pre-trained model to understand the physical flow based on changes in each variable and across vertical levels is crucial for precipitation prediction. Despite masking an overload of information in each variable, the NWP model accurately predicted prominent patterns across all variables. In particular, we observed a resemblance in the predicted patterns for the `rain' variable despite the masked pixels carrying limited information for this instance and inherent nonlinearity of the variable.

\begin{figure}[htb]
\centering
\begin{tabular}{c@{\hspace{2pt}}c@{\hspace{2pt}}c@{\hspace{2pt}}c@{\hspace{2pt}}c}
\textbf{\scriptsize (a) Ground-truth \quad} & 
\textbf{\scriptsize (b) RDAPS \quad} & 
\textbf{\scriptsize (c) Metnet \quad} &
\textbf{\scriptsize (d) Ours \quad}\\
\includegraphics[width=0.22\textwidth, height=5cm]{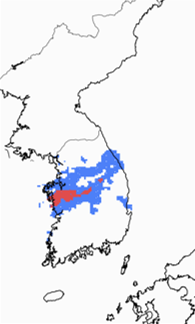} & 
\includegraphics[width=0.22\textwidth, height=5cm]{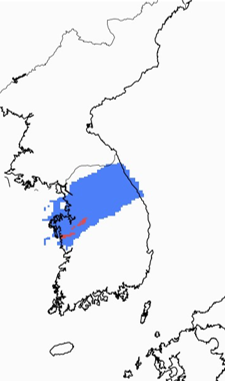} & 
\includegraphics[width=0.22\textwidth, height=5cm]{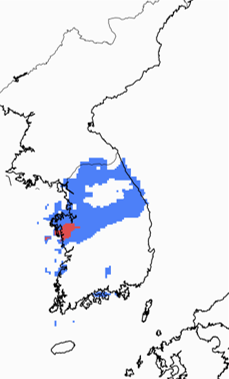} & 
\includegraphics[width=0.22\textwidth, height=5cm]{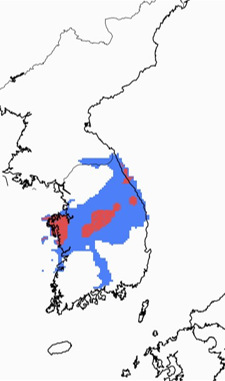} &
\includegraphics[width=0.05\textwidth, height=5cm]{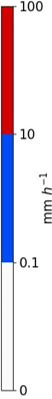}\\
\end{tabular}
\caption{Qualitative comparison between models trained on data from August 10 2022 at 00 UTC. Each result represents a cumulative result over a 1-hour period.
Owing to the influence of a stagnant front, rain fell in most parts of Korea; the average rainfall is 100$\sim$200 mm per day, and the maximum exceeds 300 mm.}
\label{fig:visual}
\end{figure}

Figure \ref{fig:visual} shows that above 10 mm of precipitation covers much of the Korean Peninsula. A deep learning-based post-processing model captures above 10 mm of rainfall, while NWP models (RDAPS) underestimate this case.
In Figure \ref{fig:visual} (c), when trained with Algorithm \ref{alg1}, Metnet exhibits errors in the rainfall location but demonstrated accurate predictions for 10 mm of rainfall.
In Figure \ref{fig:visual} (d), for the proposed model, we observed predictions for rainfall location and 10 mm of rainfall, but these tended to be overestimated.
Based on the results, combining $p(y)$ and $p^{*}(y)$ is expected to further enhance the accuracy.
The limitation lies in the fact that the input data relies on NWP predictions. 
As a result, the prediction model learns with a certain margin of error and predicts in a manner similar to RDAPS in some cases.

\section{Conclusion}
This study proposes an approach for precipitation post-processing based on transfer learning for target domain adaptation. Our main contributions involve (1) applying transfer learning to the atmosphere system by conducting pre-training on the reconstruction domain and integrating the parameters in the segmentation domain and 2) employing stochastic softening one-hot labeling to overcome biased learning from unbalanced data.
Our experiments on domain adaptation using the two-step training strategy show that the proposed method helps understand the correlation among atmospheric variables. This yields a better transfer learning performance from the precipitation post-processor.

\section{Acknowledgments}
This work was carried out through the R\&D project ``Development of a Next-Generation Operational System by the Korea Institute of Atmospheric Prediction Systems (KIAPS)", funded by the Korea Meteorological Administration (KMA2020-02213).

\bibliographystyle{plainnat}
\bibliography{bibliography}

\end{document}